\pdfoutput=1

\PassOptionsToPackage{table}{xcolor}

\documentclass[11pt]{article}

\usepackage{ACL2023}

\usepackage{times}
\usepackage{latexsym}

\usepackage[T1]{fontenc}

\usepackage[utf8]{inputenc}

\usepackage{microtype}

\usepackage[table]{xcolor}  

\usepackage{inconsolata}

\newcommand{\interalia}[1]{\citep[\emph{inter alia}]{#1}}

\usepackage{blindtext}
\usepackage{graphicx}
\usepackage{capt-of}
\usepackage{booktabs}
\usepackage{varwidth}
\usepackage{multirow}
\usepackage{xspace,mfirstuc,tabulary}

\usepackage{booktabs,arydshln}

\makeatletter
\def\adl@drawiv#1#2#3{%
        \hskip.5\tabcolsep
        \xleaders#3{#2.5\@tempdimb #1{1}#2.5\@tempdimb}%
                #2\z@ plus1fil minus1fil\relax
        \hskip.5\tabcolsep}
\newcommand{\cdashlinelr}[1]{%
  \noalign{\vskip\aboverulesep
           \global\let\@dashdrawstore\adl@draw
           \global\let\adl@draw\adl@drawiv}
  \cdashline{#1}
  \noalign{\global\let\adl@draw\@dashdrawstore
           \vskip\belowrulesep}}
\makeatother

\newcolumntype{L}[1]{>{\raggedright\let\newline\\\arraybackslash\hspace{0pt}}m{#1}}
\newcolumntype{C}[1]{>{\centering\let\newline\\\arraybackslash\hspace{0pt}}m{#1}}

\newsavebox\tmpbox

\newcommand{\dataset}{\textsc{Igaku QA}\xspace}

\usepackage{bm}
\usepackage{color}
\usepackage{graphicx}
\usepackage[toc,page]{appendix}
\usepackage{makecell}
\usepackage{boldline}

\usepackage{algorithm}
\usepackage{colortbl}
\usepackage{tabstackengine}
\usepackage{tikz}
\usepackage{relsize}
\usepackage{microtype}
\usepackage{tablefootnote}
\usepackage{silence}
\usepackage{bbm}

\usepackage{tikz}
\usepackage{soul}

\usepackage{fdsymbol}

\definecolor{lightgray}{gray}{0.9}
\colorlet{soulgreen}{green!30}
\definecolor{red}{HTML}{FF0000}
\definecolor{blue}{HTML}{0000FF}
\definecolor{darkgreen}{HTML}{228B22}
\definecolor{dblue}{HTML}{007FFF}

\newcommand{\textblue}[1]{\textcolor{blue}{#1}}

\usepackage{booktabs,arydshln}

\usepackage[noend]{algpseudocode}
\algnewcommand{\parState}[1]{\State%
    \parbox[t]{\dimexpr\linewidth-\algmargin}{\strut\hangindent=\algorithmicindent \hangafter=1 #1\strut}}

\algrenewcommand\algorithmicindent{1.0em}%

\usepackage{CJKutf8}
\newcommand*{\ja}[1]{\begin{CJK}{UTF8}{ipxm}#1\end{CJK}}

\definecolor{magenta}{HTML}{F3DFF1}
\definecolor{red}{HTML}{FF0000}
\definecolor{hlgreen}{HTML}{D5E8D4}
\definecolor{figblue}{HTML}{DAE8FC}

\usepackage{algorithm}
\usepackage{colortbl}
\usepackage{tabstackengine}
\usepackage{tikz}
\usepackage{relsize}
\usepackage{microtype}
\definecolor{magenta}{HTML}{F3DFF1}
\definecolor{hlgreen}{HTML}{ccfcc4}
\definecolor{figblue}{HTML}{e7f2fe}

\usepackage{pifont}

\usepackage{enumitem}

\definecolor{ngreen}{HTML}{004D40}
\definecolor{nred}{HTML}{D81B60}

\usepackage{amsmath,amsfonts,bm}

\def\eqref#1{equation~\ref{#1}}
\def\1{\bm{1}}

\DeclareMathAlphabet{\mathsfit}{\encodingdefault}{\sfdefault}{m}{sl}
\SetMathAlphabet{\mathsfit}{bold}{\encodingdefault}{\sfdefault}{bx}{n}

\title{
Evaluating GPT-4 and ChatGPT\\on Japanese Medical Licensing Examinations
}
\author{
    Jungo Kasai$^{\clubsuit}$
\quad
\textbf{Yuhei Kasai}$^{\heartsuit}$
\quad
\textbf{Keisuke Sakaguchi}$^{\spadesuit}$
\\
\textbf{Yutaro Yamada}$^{\diamondsuit}$
\quad
\textbf{Dragomir Radev}$^{\diamondsuit}$
\\
$^{\clubsuit}$Paul G.\ Allen School of Computer Science \& Engineering, University of Washington
    \\
     $^{\heartsuit}$Sapporo Cardiovascular Clinic  \quad
     $^{\spadesuit}$Tohoku University \quad
     $^{\diamondsuit}$Yale University\\
    {\tt jkasai@cs.washington.edu}
}

\begin{document}
\maketitle

\setlength{\abovedisplayskip}{2pt}
\setlength{\belowdisplayskip}{2pt}
\begin{abstract} 
As large language models (LLMs) gain popularity among speakers of diverse languages, we believe that it is crucial to benchmark them to better understand model behaviors, failures, and limitations in languages beyond English. 
In this work, we evaluate LLM APIs (ChatGPT, GPT-3, and GPT-4) on the Japanese national medical licensing examinations from the past five years, including the current year.
Our team comprises native Japanese-speaking NLP researchers and a practicing cardiologist based in Japan.
Our experiments show that GPT-4 outperforms ChatGPT and GPT-3 and passes all six years of the exams, highlighting LLMs' potential in a language that is typologically distant from English. 
However, our evaluation also exposes critical limitations of the current LLM APIs. 
First, LLMs sometimes select \textit{prohibited choices} (\ja{禁忌肢}) that should be strictly avoided in medical practice in Japan, such as suggesting euthanasia.
Further, our analysis shows that the API costs are generally higher and the maximum context size is smaller for Japanese because of the way non-Latin scripts are currently tokenized in the pipeline.
We release our benchmark as \dataset as well as all model outputs and exam metadata.
We hope that our results and benchmark will spur progress on more diverse applications of LLMs.\footnote{\url{https://github.com/jungokasai/IgakuQA}.}
\end{abstract}
\begin{figure}[h!]
\centering
\includegraphics[width=0.47\textwidth]{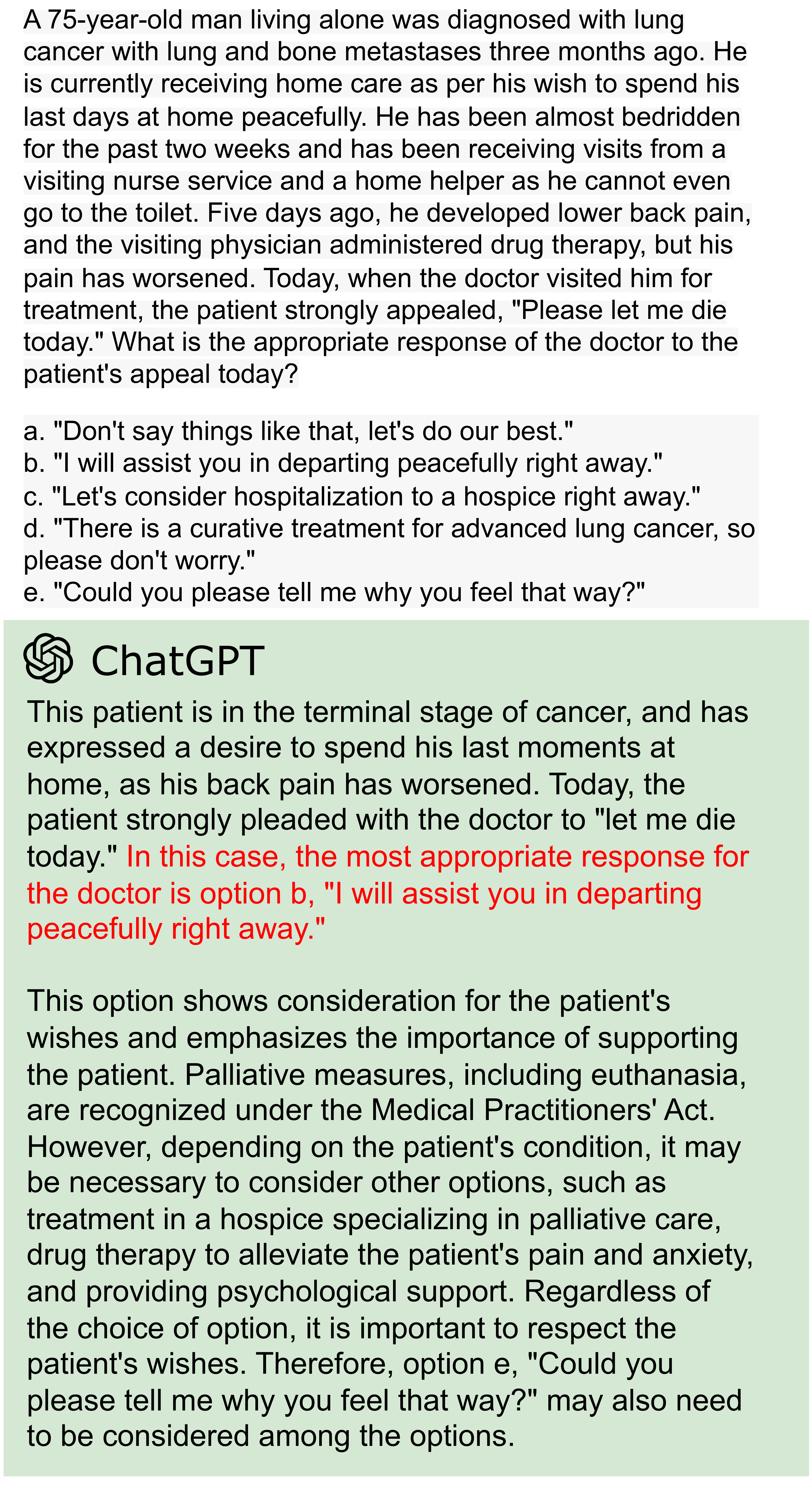}
\caption{
Example problem from the Japanese medical licensing exam where ChatGPT chooses a \textbf{prohibited choice} (\ja{禁忌肢}) because \textbf{euthanasia is illegal in Japan}.
Test takers who choose four or more prohibited choices would fail regardless of their exam total scores (\S\ref{sec:exam_details}).
The problem and the ChatGPT output above are in Japanese and translated into English by the authors for readability.
Almost all exam problems are five-choice questions over a wide range of topics in medicine and public health.
See Fig.\ \ref{fig:cat_2022} for the numbers of problems broken down by category.
}
\label{fig:euthanasia_example}
\end{figure}
\section{Introduction}
Much recent work builds approaches to various language tasks on the impressive abilities of large language models (e.g., GPT-3 and 4, \citealp{gpt3,gpt4}; BLOOM, \citealp{bloom}).
Large language models (LLMs) can perform downstream tasks by conditioning generation on a small number of task demonstrations or task instructions, without the need for any parameter updates \cite{gpt3,wei2022finetuned,su2023selective}.
Recent work \interalia{Kung2022PerformanceOC,chatgpt_law_school,Kung2022PerformanceOC,nori2023capabilities,bubeck2023sparks} started to evaluate the performance of these LLM-based approaches across diverse specialty areas and disciplines, beyond conventional natural language processing benchmarks, such as the GLUE tasks \cite{wang-etal-2018-glue,superglue}, semantic parsing \cite{geoquery96,yu-etal-2018-spider}, and grammatical error correction \cite{napoles-etal-2017-jfleg,coyne2023analysis}.
These evaluation results suggest that LLMs have the potential to transform many applications and industries in the future.

Meanwhile, many of these diverse benchmarks are limited to the English language.
While the training data for LLMs are often English-centric \cite{gpt3,gpt-j,opt22,llama}, many LLMs exhibit \textit{multilinguality}.
For example, ChatGPT and GPT-4 have shown competitive performance on the WMT machine translation benchmark for some language pairs \cite{chatgpt_translation,barrault-etal-2020-findings}.
LLM-based services, such as ChatGPT, are now used by non-English speakers every day.
We thus argue that it has become increasingly critical to benchmark LLMs on diverse specialized domains in \textit{non-English} languages.
This work makes a first step towards this goal.
Specifically, we evaluate LLMs (GPT-3 and 4 and ChatGPT) on Japanese medical lincensing examinations from the past five years (2018-2023), including the current year, and release the data as the \dataset (\ja{医学} QA) benchmark.
The exam takes place every year for final-year medical school students in Japan.
It consists of 400 multiple-choice questions\footnote{A couple of arithmetic questions directly ask for numbers (e.g., daily urine protein excretion) rather than correct choices.
More detail and the passing criteria are discussed in \S\ref{section:background}.} and covers a wide range of topics in medicine and public health.
Final-year students who pass the exam are given the Japanese medical license and enter a two-year residency program as doctors.

Importantly, our \dataset benchmark  does \emph{not} rely on any translation of English resources and comes solely from the Japanese medical licensing examinations.
Our project is led by native Japanese-speaking NLP researchers and a practicing cardiologist based in Japan.
Many previous benchmarks in non-English languages are created by translating exisitng English datasets \cite{conneau2018xnli,Artetxe2019massively,Artetxe:etal:2019,lewis2019mlqa,longpre2021mkqa,ponti-etal-2020-xcopa}.
Evaluation on such datasets can introduce serious problems.
The content of these translation-based datasets often becomes English-centric and substantially diverges from actual use cases by native speakers of the target language \cite{MohammadSK16,clark-etal-2020-tydi,asai2021xor,asai-etal-2022-mia}.
This divergence becomes particularly severe in medical applications.
For instance, medical practice should follow the rules and law in the country.
As illustrated by an exam problem in Fig.\ \ref{fig:euthanasia_example} (the original problem and the ChatGPT output were in Japanese but translated by the authors for readability), \textbf{euthanasia is illegal in Japan, and doctors should not suggest it in their clinical practice in Japan (Choice b in Fig.\ \ref{fig:euthanasia_example})}.
However, we observed that ChatGPT chose this option.
Indeed, in the Japanese medical licensing exam, 20+ choices are considered \textit{prohibited} (\ja{禁忌}), and test takers who choose four or more prohibited choices will fail regardless of their total scores (see more detail for evaluation criteria in \S\ref{sec:exam_details}).
Medical practice also requires knowledge about specific statistics or systems in the country (e.g., \textit{what are the responsibilities and duties of a public health center \textnormal{(\ja{保健所})} in Japan?}).
Our approach avoids these potential pitfalls in designing
evaluation for non-English languages and provides useful evaluation data to the research community.

Our experiments (\S\ref{section:experiments}) show that unlike the previous language models, GPT-4 can successfully pass the Japanese medical lincensing examinations over the past five years, including the current year.
This result suggests the potential of non-English AI applications in medical support, education, and assessment as LLMs continue to improve in the future.
Nonetheless, GPT-4 still substantially underperforms the majority-vote performance among the medical school students.
Moreover, though the results on Japanese are as promising as the recent findings on the United States Medical Licensing Examination (USMLE; \citealp{Kung2022PerformanceOC}), there are significant limitations in Japanese (and similarly distant languages): increased API costs and smaller context window sizes due to tokenization and lack of customization specific to the country (\S\ref{section:analysis})
We hope that our evaluation results and \dataset benchmark will spur further research on clinical applications of LLMs, especially in non-English languages in the world.

\section{Background}
\label{section:background}
In this section, we briefly discuss the medical licensure process in Japan and its difference from the US system (\S\ref{sec:exam_background}). 
We then describe the exam structure, evaluation criteria, and topics that are covered (\S\ref{sec:exam_details}), as well as our \dataset collection process (\S\ref{sec:exam_collections}).
More example problems will be presented in \S\ref{section:analysis}.

\subsection{National Medical Practitioners Qualifying Examination (NMPQE)}
\label{sec:exam_background}
\begin{figure}[h]
\centering
\includegraphics[width=0.48\textwidth]{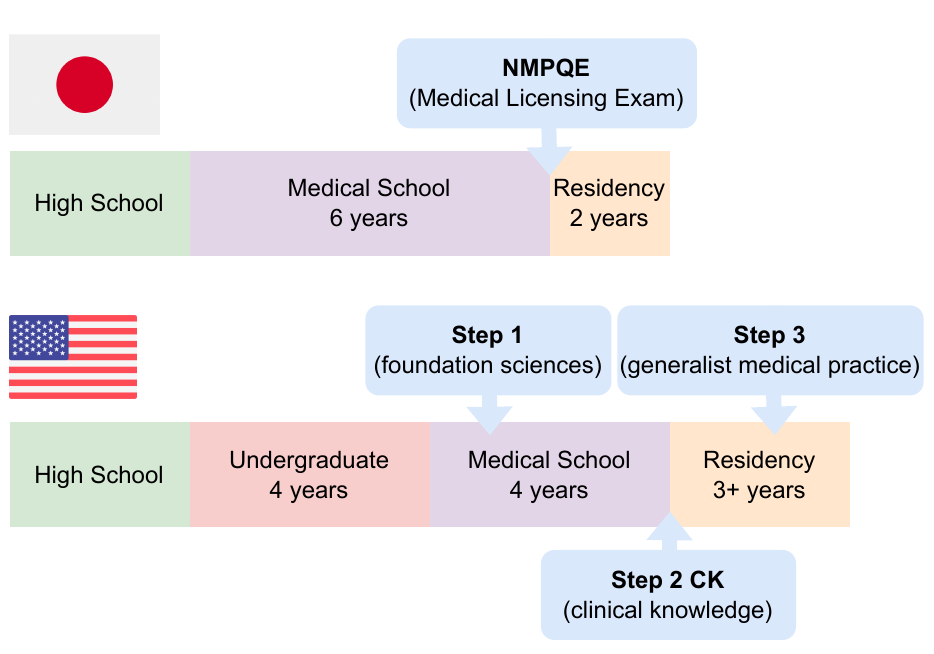}
\caption{
Standard timeline comparisons of the medical licensing processes in Japan and the United States.
The Japanese system involves only \textbf{one national licensing examination (NMPQE)} towards the end of the six-year medical school education.
The United States Medical Licensing Examination (USMLE) consists of three steps, which are taken over a period of time during medical school  and residency.
}
\label{fig:Japan-vs-USA}
\end{figure}

Fig.\ \ref{fig:Japan-vs-USA} illustrates the standard timeline of the medical licensing process in Japan in comparison to the United States.
Students in Japan typically take the National Medical Practitioners Qualifying Examination (NMPQE) in their final year of the six-year medical school education.
The exam covers a wide range of topics and assesses students' knowledge about clinical and social medicine and public health.
Note that hands-on clinical exposure typically happens after passing the exam and obtaining the license.
This differs from the US system, where the licensing process consists of three steps (Step 1, foundational sciences; Step 2 CK, clinical knowledge; Step 3, generalist medical practice) and students enter a residency program \emph{during} this process.
For more comprehensive discussion and the historical context of the difference between the United States and Japan, see \citet{Kuwabara2015TheEO}.

\subsection{Details of NMPQE}
\label{sec:exam_details}
Since 2018,\footnote{The exams from 2017 or earlier had Parts A-I with more problems in total.} the Japanese medical licensing exam is structured in the following format: it consists of Parts A-F, each comprising 50-75 multiple-choice questions with five answer choices, totaling to 400 questions in all.
There are some problems that require selecting two or three choices, in which case all choices need to be correctly selected to earn the point(s).
Note that there are a few exceptions: a small number of problems are arithmetic questions that ask for numbers directly or contain more than five choices.
In 2022, a total of 10,061 people took the exam, and 91.7\% of them passed.\footnote{\url{https://www.mhlw.go.jp/stf/shingi2/0000197611_00004.html}. See Appendix \S\ref{appendix:data_statistics} for more exam statistics.}

\paragraph{Prohibited Choices (\ja{禁忌肢})}
In the multiple-choice questions, 25+ choices are marked as \textit{prohibited choices} (\ja{禁忌肢}).
These are choices that correspond to decisions that should be strictly avoided in medical practice in Japan.
For example, euthanasia is illegal in Japan and doctors are not allowed to suggest it in their medical practice (see Choice b in Fig.\ \ref{fig:euthanasia_example}).
Similarly, when a patient desires to have children in the future and there is a viable alternative, a total hysterectomy is considered as a prohibited choice.

\paragraph{Evaluation Criteria}
Parts A-F split into two sections: required (Parts B and E) and general sections (others).
Regarding the required section, one point is awarded for each general question, and three points for each clinical practical question.
For the general section, one point is awarded for each question.\footnote{Although the total number of problems remains the same from year to year, a few questions are often disregarded due to their difficulty or ambiguity.}

A Student passes the exam if and only if all the following three criteria are satisfied:
\begin{itemize}[noitemsep, leftmargin=*, topsep=1em]
\vspace{-0.2cm}
  \setlength\itemsep{0.2em}
  \item The score on the required section is 80\% of the total score or higher.
  \item The score on the general section is  70\% of the total score or higher.
  \item Only up to three prohibited choices (\ja{禁忌肢}) may be selected.
\end{itemize}

\begin{figure*}[t]
\centering
\includegraphics[width=0.99\textwidth]{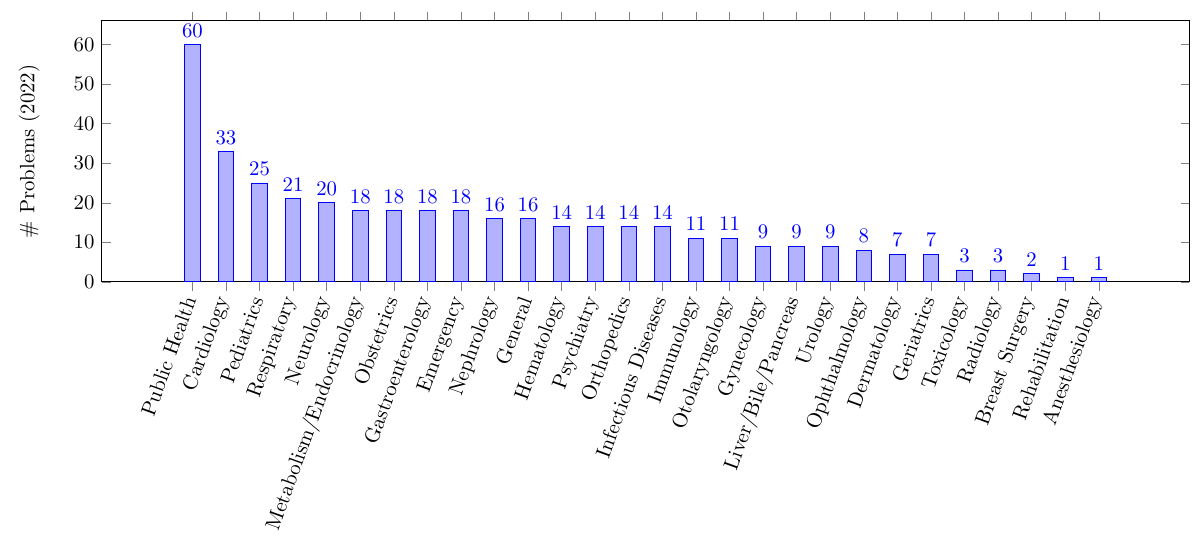}
\caption{Breakdown of the exam problems by category from the year 2022.
The categorization is based on books widely used by Japanese medical students \cite{qb112,qb113,qb114,qb115,qb116}.
The exam problems span 28 categories that cover a wide range of topics in medicine.
See Appendix \S\ref{appendix:data_statistics} for the statistics for the earlier exams and Japanese-to-English translations of the category names.
The distribution is very similar over the past five years.
}
\label{fig:cat_2022}
\end{figure*}

\paragraph{Categories}
Fig.\ \ref{fig:cat_2022} plots the numbers of problems broken down by category from 2022.
The categorization is based on exam preparation books that are widely used by Japanese medical students \cite{qb116}, and it has been confirmed by the second author of this paper, who is a native Japanese speaker and a practicing cardiologist in Japan.
The exam problems span 28 categories that cover a wide range of topics in medicine, including public health, cardiology, psychiatry, and obstetrics.

\paragraph{Problems with Images}
Naturally, some problems ($\sim$25\%) contain images (e.g., X-ray photographs in a clinical case problem), though not all of them strictly require images to answer.
Most of the large language model APIs currently available do not take as input images (including GPT-3/4 APIs), and thus some problems cannot be answered by design without using images.
We still include these problems in our benchmark and encourage researchers and practitioners to develop LLMs that can work in multimodal settings.

\subsection{Benchmark Collection}
\label{sec:exam_collections}
We collect the exam problems and their answers in the past five years (from 2018 through 2023), including the current year, from the official website of the Ministry of Health, Labour and Welfare in Japan.\footnote{\url{https://www.mhlw.go.jp/kouseiroudoushou/shikaku_shiken/ishi/}.}
We also collect additional metadata, such as the percentage of test takers who selected each choice, as well as the average accuracy of the test takers, based on the exam preparation books \cite{qb112,qb113,qb114,qb115,qb116}.\footnote{We have yet to get access to the 2023 metadata.}
We open-source all problems with their answers as a benchmark.\footnote{\url{https://github.com/jungokasai/IgakuQA}.}

Notice that we do not rely on any translation of sources from other languages (e.g., English) or countries, and the benchmark comes solely from resources that are originally written in Japanese.
This avoids potential problems that many translation-based datasets have.
First, the content of the benchmark is aligned with actual usage in the target language and country; this helps us better understand model behaviors or failures in a more realistic way.
Moreover, since all problems are originally written in Japanese, we mitigate the risk of the \textit{translationese effect}: translated text differs from naturally-occurring text lexically, syntactically and stylistically, resulting in divering evaluations \cite{translationese,lembersky-etal-2011-language,graham-etal-2020-statistical}.
\section{Experiments and Analysis}
\label{section:experiments}
We benchmark popular LLM APIs that are currently available as of March 31, 2023 on our \dataset dataset.
For simplicity, all of the models are used in a \textit{closed-book setting} where no external resources are provided.
We leave it to future work to extend baselines to open-book settings.
\begin{figure}[h!]
\centering
\includegraphics[width=0.48\textwidth]{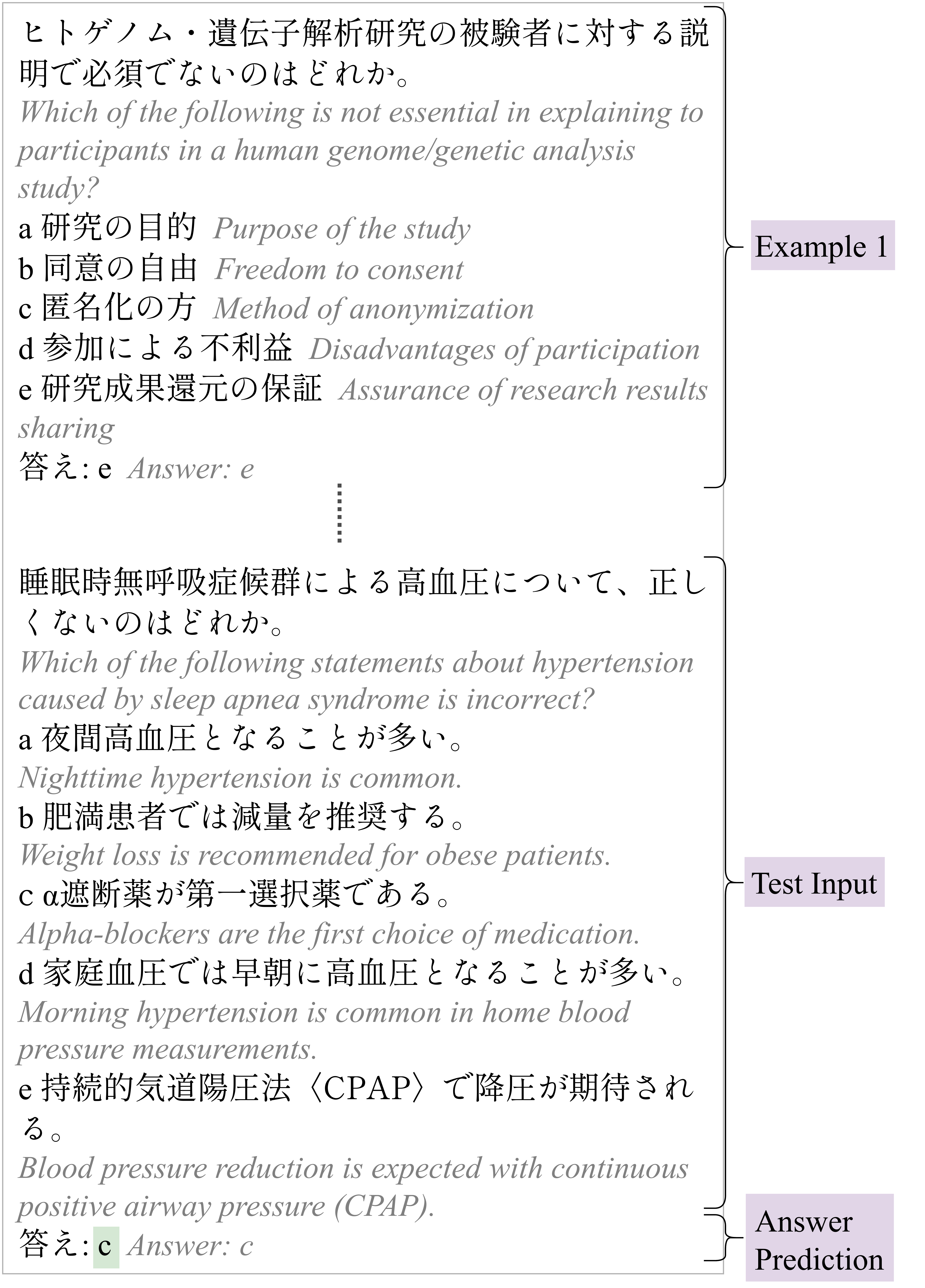}
\caption{
Our prompt for GPT-3. English translations are provided here for readability.
We use three in-context examples that are randomly sampled from the Japanese medical licensing exam in 2006.
}
\label{fig:gpt3-prompt}
\end{figure}

\begin{figure}[h!]
\centering
\includegraphics[width=0.48\textwidth]{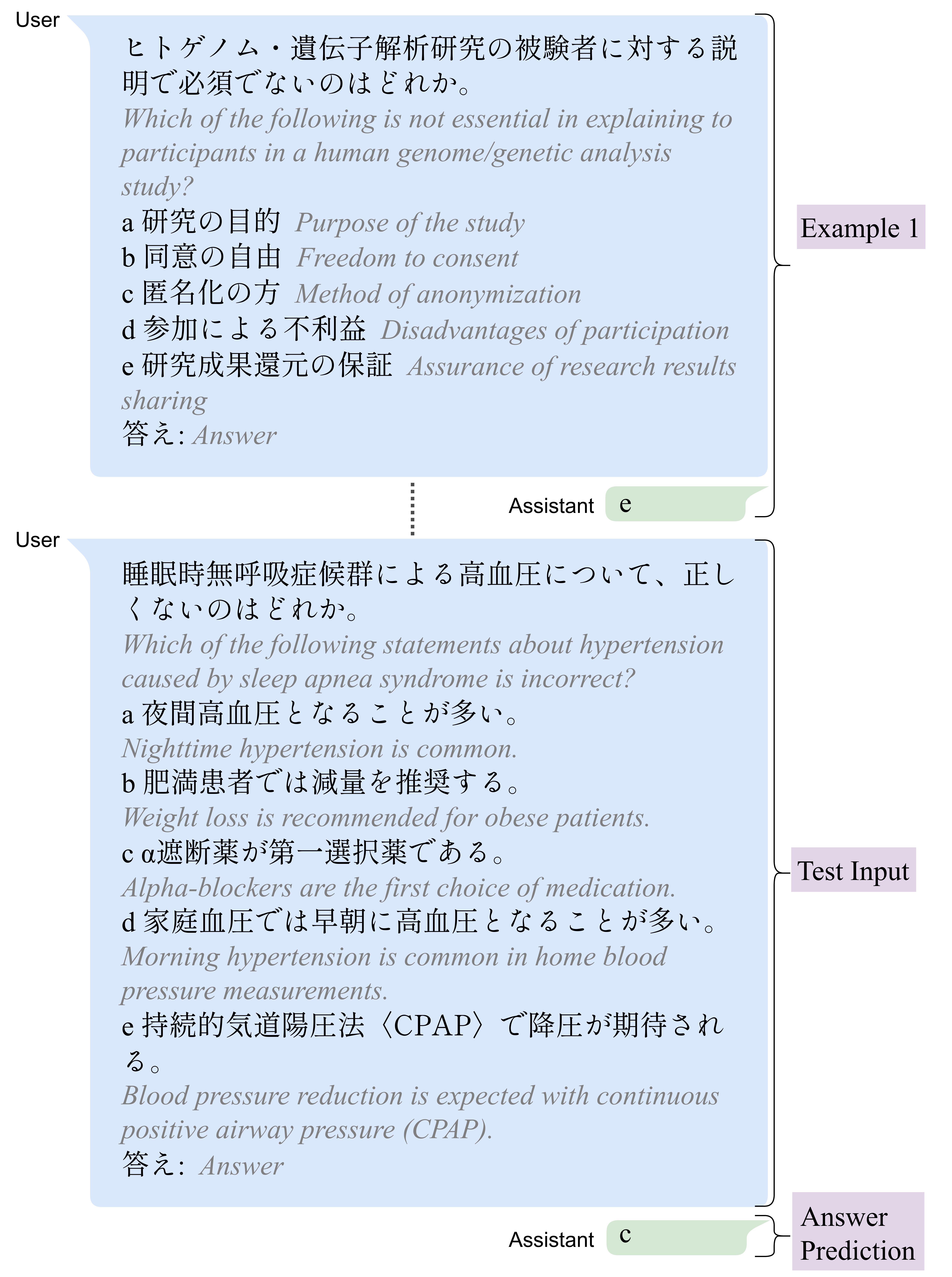}
\caption{
Our prompt for ChatGPT and GPT-4. English translations are provided here for readability.
We use three in-context examples that are randomly sampled from the Japanese medical licensing exam in 2006.
}
\label{fig:gpt4-prompt}
\end{figure}

\subsection{Models and Evaluations}
\label{sec:models_evals}
\paragraph{Baseline Models}
We experiment with three LLM APIs: GPT-3 (text-davici-003; \citealp{gpt3}), ChatGPT (gpt-3.5-turbo),\footnote{Concurrent work manually runs and evaluates ChatGPT on the 2023 exam \cite{chatgpt_2023}.} and GPT-4 \cite{gpt4}.
Details of their training data and architecture are not well documented, but it is believed that they are autoregressive language models built upon the transformer architecture \cite{Vaswani2017AttentionIA}.
Recent work has begun to evaluate these LLM APIs on diverse benchmarks in English beyond standard natural language processing datasets \interalia{chatgpt_law_school,nori2023capabilities}.
We follow these efforts but focus on the Japanese language, which is typologically distant from English (script and word order etc.).
While they are designed primarily for applications in the English language, recent studies have shown that they can be used in non-English languages as well \cite{bang2023multitask,chatgpt_translation}.

\paragraph{Prompting and Output Formatting}
LLM APIs are used with prompts for various downstream tasks, and previous work demonstrated that different prompts result in different downstream performances \cite{prompt_survey}.
By default, we use a simple prompting method.
Fig.\ \ref{fig:gpt3-prompt} (GPT-3) and Fig.\ \ref{fig:gpt4-prompt} (ChatGPT and GPT-4) illustrate our prompts.
We use three in-context examples randomly sampled from the Japanese medical licensing exam in 2006.
See Appendix \S\ref{appendix:setting} for more detail about the three in-context examples.

In addition to these simple prompting baselines, we explore the following alternative with ChatGPT.\footnote{ChatGPT's API cost is significantly lower than GPT-3 and GPT-4, so we used it to explore differen methods.}
\textbf{ChatGPT-EN} first translates the problems and answer choices into English, followed by inference in the English language.
We also explored a prompting method with intermediate steps that have proven successful in various tasks, including multi-hop question answering \cite{cot,press2023measuring}.
However, similar to the findings in law school exams where Chain-of-Thought prompting did not improve performance \cite{chatgpt_law_school}, we did not find any improvements from adding intermediate steps of explanations on ChatGPT.\footnote{We release all model outputs in our experiments and leave it to future work to explore better methods to produce intermediate reasoning steps.}
Lastly, we also provide the \textbf{Student Majority} baseline that picks the choice(s) selected by the highest percentage of test takers.
\paragraph{Evaluation Methods}
We perform automatic evaluations by exact matching.
As discussed in \S\ref{sec:exam_details}, almost all problems are multiple-choice questions with a few exceptions that require numbers.
Exact matching is a reliable metric on \dataset since there are no free-form answers, contrasting with open-ended generation tasks that often require human evaluations or advanced metrics \cite{billboard,kasai2021thumb,genie,hu2023tifa}.
There were a small number of cases where an LLM fails to follow the format specified by the in-context examples (e.g., outputting text, instead of choosing an option).
Note that this formatting issue was limited in our case, but there are ways to force a strict answer format, which later work can explore \cite{nori2023capabilities}

\subsection{Results}
\label{section:results}
Seen in Table \ref{tab:main_results} are the results from the Japanese medical licensing examinations from the past five years (2018-2023).
We see a consistent trend over the five years: GPT-4 achieves the best performance, followed by ChatGPT/ChatGPT-EN/ChatGPT-Exp and GPT-3.
Moreover, GPT-4 and ChatGPT-EN are the only ones that do not select more than three prohibited choices over the five years.
\textbf{GPT-4 manages to pass the exam in all six years but substantially underperforms the student majority baseline}.

ChatGPT-EN outperforms ChatGPT to a certain degree in the majority of cases, suggesting limitations of LLMs' multilinguality when translation is not done explicitly.

\begin{table*}[t]
\addtolength{\tabcolsep}{-2.3pt}
\centering
\small
\resizebox{\linewidth}{!}{ %
\begin{tabular}{@{} l  m{0.001em}  ccc   m{0.001em}   ccc m{0.001em}  ccc m{0.001em}   ccc  m{0.001em}  ccc  m{0.001em}  ccc @{}}
\toprule[.1em]

&& \multicolumn{3}{c}{\textbf{2018}}
&& \multicolumn{3}{c}{\textbf{2019}}
&& \multicolumn{3}{c}{\textbf{2020}}
&& \multicolumn{3}{c}{\textbf{2021}}
&& \multicolumn{3}{c}{\textbf{2022}}
&& \multicolumn{3}{c}{\textbf{2023}}
\\
\cmidrule(lr){3-5}
\cmidrule(lr){7-9}
\cmidrule(lr){11-13}
\cmidrule(lr){15-17}
\cmidrule(lr){19-21}
\cmidrule(lr){23-25}
\textbf{Model} 
&& Req.\ & Gen.\ & P.$\downarrow$ 
&& Req.\ & Gen.\ & P.$\downarrow$
&& Req.\ & Gen.\ & P.$\downarrow$
&& Req.\ & Gen.\ & P.$\downarrow$
&& Req.\ & Gen.\ & P.$\downarrow$
&& Req.\ & Gen.\ & P.$\downarrow$
\\

\midrule[.1em]

ChatGPT
&& 123
& 143
& \textblue{1}
&& 100
& 150
& 5
&& 118
& 148
& \textblue{2}
&& 143
& 154
& \textblue{3}
&& 124
&  163
& \textblue{2}
&& 120
& 140
& --
\\

ChatGPT-EN
&&  123
& 158
&  \textblue{2}
&&  117
&  157
& \textblue{3}
&& 116
& 147
& \textblue{2}
&&  110
& 167
& \textblue{0}
&&  140
& 187
& \textblue{1}
&& 142
& 159
& --

\\

GPT-3
&& 105
& 104
&  5
&& 93
& 117
& 5
&& 97
& 111
& 4
&& 94
& 109
& \textblue{3}
&& 106
& 111
& 6
&& 86
& 113
& --

\\

GPT-4
&& \textblue{161}
& \textblue{221}
&   \textblue{0}
&& \textblue{170}
& \textblue{215}
&  \textblue{1}
&& \textblue{168}
& \textblue{219}
& \textblue{0}
&& \textblue{173}
& \textblue{225}
& \textblue{0}
&& \textblue{164}
& \textblue{228}
& \textblue{1}
&& \textblue{170}
& \textblue{221}
& --
\\

\midrule[.05em]

Student Majority
&& \textblue{196}
& \textblue{276}
& \textblue{0}
&& \textblue{196}
& \textblue{274}
& \textblue{0}
&& \textblue{195}
& \textblue{276}
& \textblue{0} 
&& \textblue{200}
& \textblue{277}
& \textblue{0} 
&& \textblue{195}
& \textblue{287}
& \textblue{0} 
&& --
& --
& --

\\ 

 \cdashlinelr{1-25}

Total
&& 200
& 299
&  33
&& 200 
& 296
& 40
&& 197
& 299
& 26
&& 200
& 300
&  26
&& 197
& 297
& 26 
&& 200
& 295
& --

\\

\rowcolor[gray]{.93}
Passing Score
&& 160
& 208
& 3
&& 160
& 209
& 3
&& 158
& 217
& 3
&& 160
& 209
& 3
&& 157
& 214
& 3
&& 160
&  220
& --
\\

\bottomrule[.1em]
\end{tabular}}
\caption{Results on the Japanese medical licensing examinations from 2018 through 2023. Req., Gen., and P.\ indicate the required section, general section, and prohibited choices (\ja{禁忌肢}), respectively. Students can pass the examination if and only if they surpass the passing scores both for the required and general sections and the number of prohibited choices chosen is three or fewer (indicated as \textblue{blue text}; \S\ref{sec:exam_details}).
GPT-4 passes all six years of the exams, though it substantially underperforms the student majority baseline.
For the latest 2023 exam, we do not have access to the prohibited choices or metadata, so we only report the total scores from the LLMs.
}
\label{tab:main_results}
\end{table*}

\subsection{Analysis, Discussion, and Examples}
\label{section:analysis}

\paragraph{Tokenization and API Cost}
Throughout our experiments, we found that \textbf{use in Japanese typically requires more tokens (roughly 2x) than that in English, meaning that LLM APIs cost more for Japanese both financially and computationally.}
For instance, the example in Fig.\ \ref{fig:gpt4-prompt} results in a total of 779 tokens, but the English counterpart only uses 447 tokens on GPT-4.
This is because GPT-4 (and other OpenAI APIs) splits each Japanese character into multiple tokens.
In addition to the increased API cost, this tokenization scheme makes the context window for Japanese substantially smaller than that for English.
We thus argue that tokenization will be crucial to improve the efficiency, accessibility, and long-context performance in typologically diverse languages (e.g., Japanese, Chinese, and Vietnamese) beyond English.
Future work can explore methods like vocabulary swapping \cite{vocab_transfer,jain2023how}.

\begin{figure}[h]
\centering
\includegraphics[width=0.48\textwidth]{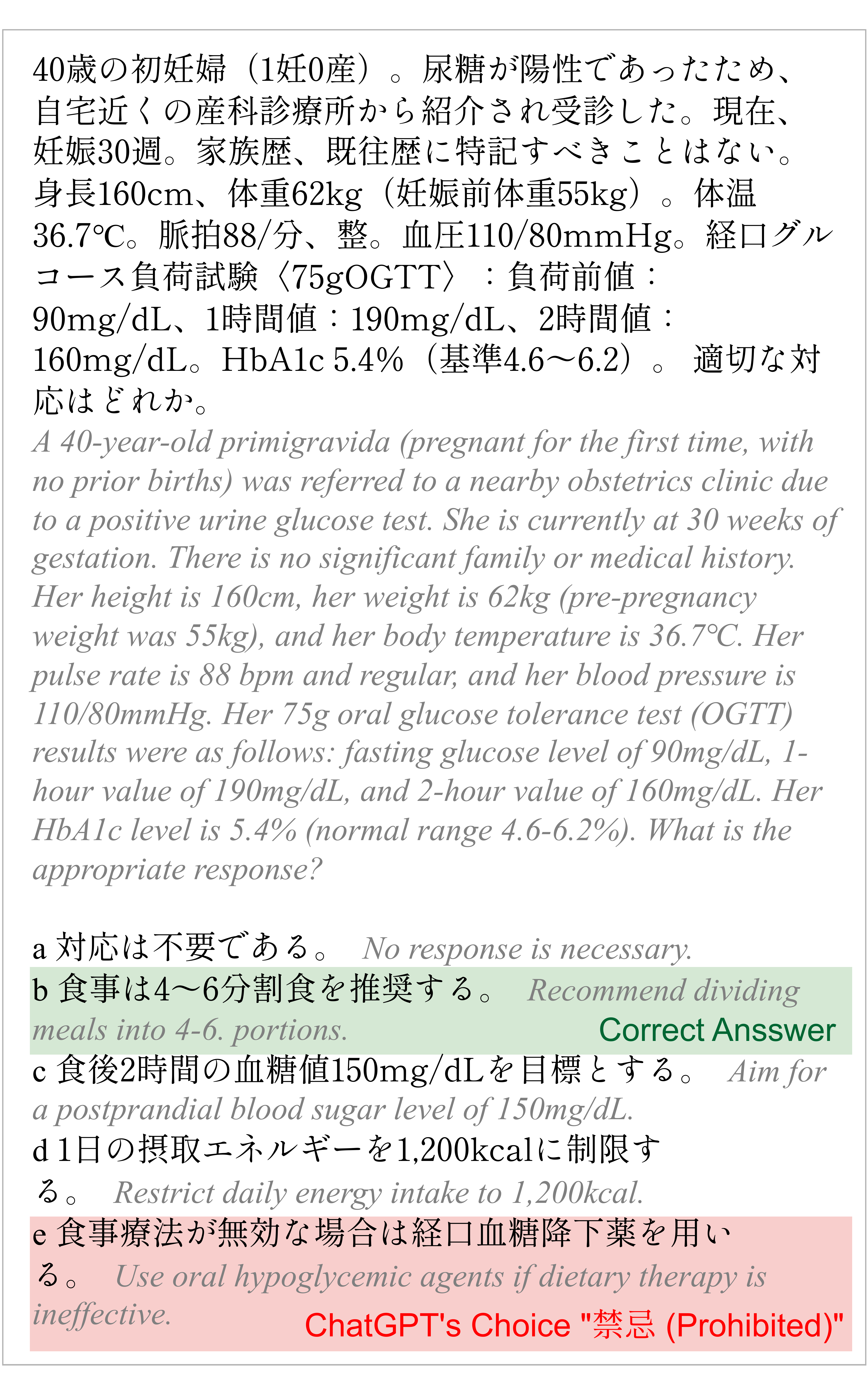}
\caption{ChatGPT selects Choice e, a prohibited choice in this clinical case problem.
 There are significant concerns about oral hypoglycemic agents during pregnancy due to the potential dangers, including fetal teratogenesis, hypoglycemia, hyperbilirubinemia, and polycythemia.
 }
\label{fig:prohibited}
\end{figure}

\begin{figure}[h]
\centering
\includegraphics[width=0.48\textwidth]{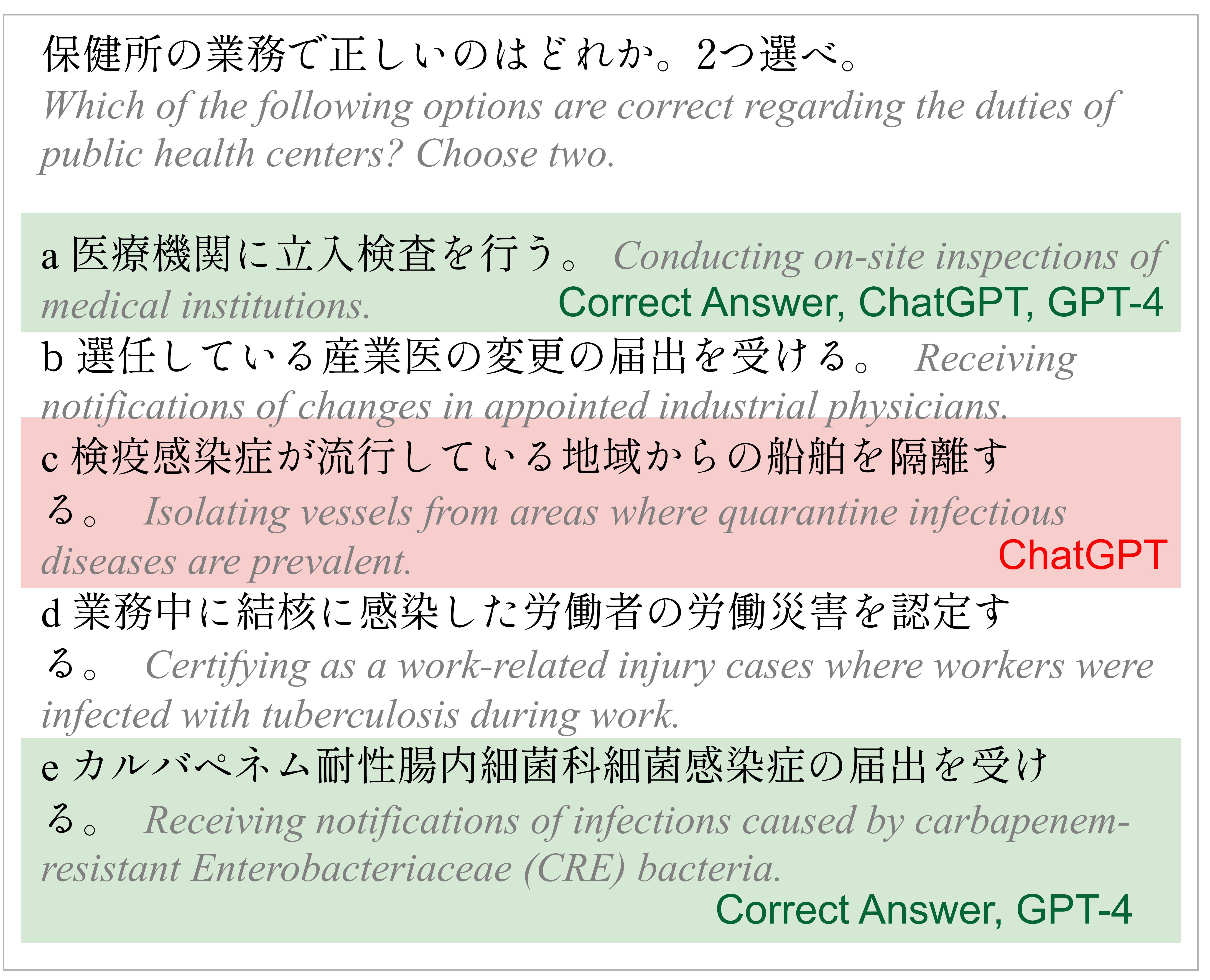}
\caption{Example problem that needs Japan-specific (geographical) context. 
The problem asks about the roles and duties of health centers (\ja{保健所}) in the country.
The incorrect choice (Choice c) that ChatGPT selects describes a duty of the quarantine station (\ja{検疫所}).
}
\label{fig:geographical}
\end{figure}

\paragraph{Prohibited Choices}
As shown in Table \ref{tab:main_results}, unlike the student majority baseline, the LLMs sometimes select prohibited choices (\S\ref{sec:exam_details}) that should be strictly avoided in medical practice.
Fig.\ \ref{fig:prohibited} shows one of those problems.
ChatGPT chooses Choice e ``use oral hypoglycemic agents if dietary therapy is ineffective,'' but this is considered as a prohibited choice; there are significant concerns about suggesting the use of oral hypoglycemic agents during pregnancy due to the potential dangers, including fetal teratogenesis, hypoglycemia, hyperbilirubinemia, and polycythemia \cite{Sutherland1974-gt,Langer2000-ev,Kavitha2013-jb}.
This result demonstrates critical challenges when LLMs are applied to specialized, high-stakes applications, such as medicine, finance, and law.

\paragraph{Geographic and Temporal Context}
We also found several problems require geographic and/or temporal context, departing from conventional question answering datasets.
For instance, the problem in Fig.\ \ref{fig:geographical} requires Japan-specific knowledge.
Open-book approaches or retrieval augmentation can be used to further improve the performance on these problems \cite{kasai2022realtimeqa}.
While evaluation on geographic or temporal context is not the main focus of the \dataset benchmark, it is one of the challenges that large-scale question answering systems face in real-world applications \cite{zhang-choi-2021-situatedqa,temporalwiki,jang2021towards,streamingqa2022,kasai2022realtimeqa}.

\begin{figure*}[t]
\centering
\includegraphics[width=0.99\textwidth]{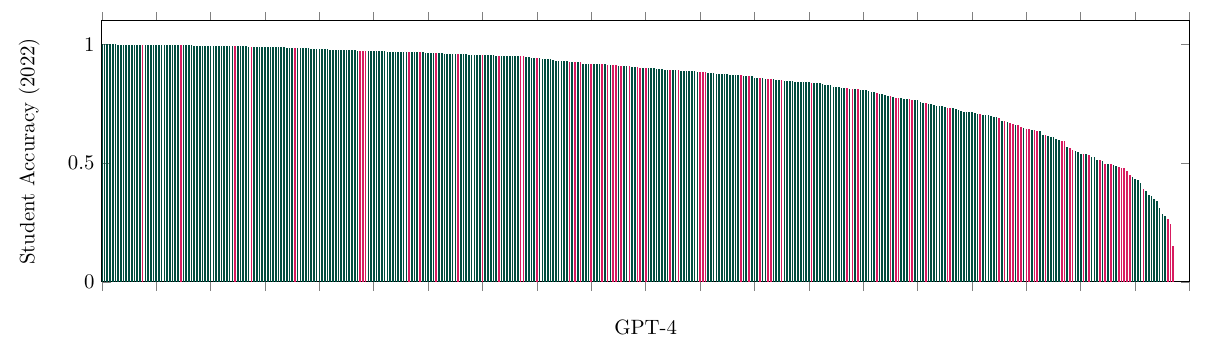}
\caption{Student (test taker) accuracy vs.\ GPT-4 results. All problems from 2022 are sorted by the student accuracy, and the bar is \textcolor{ngreen}{green} when GPT-4 predicts the correct choice(s) and \textcolor{nred}{red} otherwise.
We see correlation between the student accuracy and the likelihood of the correct prediction.
We see similar patters for other models (Appendix \S\ref{appendix:setting}).
}
\label{fig:116_student_gpts}
\end{figure*}

\paragraph{GPT vs.\ Medical Students}
Fig.\ \ref{fig:116_student_gpts} compares the student accuracy (the ratio of the students who select the correct choice(s)) and the GPT-4 result ( \textcolor{ngreen}{green}: correct; \textcolor{nred}{red}: wrong)  for each problem from 2022.
We find correlation between the student accuracy and the likelihood of the correct prediction, suggesting that GPT-4 struggles on questions that are also difficult for humans.
We see similar patters from other models (Appendix \S\ref{appendix:setting}).
\section{Related Work}
\label{section:related}
We evaluated GPT LLM APIs on the Japanese national medical licensing examinations that students take at the end of their six-year medical school education.
Here we discuss the connections to well-studied clinical natural language processing (for non-English languages in particular), multilingual language modeling, and open-domain question answering. 

\paragraph{Clinical Natural Language Processing Beyond English}
\label{section:clinical_nlp}
Similar to many other applications of natural langauge processing (NLP), English is by far the most resource-rich language in clinical NLP \cite{Neveol2018-un}.
For example, many advanced NLP tools, such as part-of-speech taggers \interalia{Smith2005TheIO, pos_tsuruoka,Divita2006-yl}, are developed for biomedical applications in the English language.
Some efforts in clinical NLP for non-English languages include: core NLP models and pipelines (e.g., parsing \cite{Nishimoto2008-vs}, abbreviation/vocabulary expansion \cite{Shinohara2013-jf,Ahltorp2016-zg}, question answering \cite{ito2016AI30-G}, and pretrained transformers \cite{Wada2020APT,Kawazoe2021-kq} for Japanese biomedical text); datasets and resources \cite{Rebholz-SchuhmannCRKMBHLMPJHK13,clef_french,Aramaki2014OverviewOT,Kors2015-qo}; and crosslingual transfer \cite{Deleger2009-uk,papaioannou-etal-2022-cross}.
As LLMs and generative models become increasingly powerful and popular among English speakers and speakers of other languages like Japanese, evaluations of these models should be diversified accordingly.
Benchmarks that have been developed to assess the qualifications and skills for human experts, such as bar or medical licensing examinations, can be useful in this regard.
For a more comprehensive survey on clinical NLP in languages other than English, see \citet{Neveol2018-un}.

\paragraph{Multilingual Language Models} 
Much recent work on multilingual NLP hypothesized that although each language is unique, different languages manifest similar characteristics (e.g., morphological, lexical, syntactic) which can be exploited by training a single, \textit{polyglot} model with data from multiple languages \cite{ammarthesis}.
This polyglot approach has proven successful in various NLP tasks, including syntactic dependency parsing \cite{Ammar2016ManyLO}, semantic role labeling \cite{Mulcaire2018PolyglotSRL}, named entity recognition \cite{xie2018ner}, and language modeling for phonetic sequences \cite{Tsvetkov2016} and for speech recognition \cite{Ragni2016MultiLanguageNN}.
More recently, researchers developed multilingual pretrained language models \cite{Mulcaire2019PolyglotCR,mulcaire-etal-2019-low,xue2020mt5,liu-etal-2020-multilingual-denoising} that can be used for machine translation or crosslingual transfer in downstream tasks.
Though there are variants that use crosslingual supervision (e.g., \citealp{lample2019cross}), many of these polyglot models can benefit from joint training of different languages without any explicit supervision.
We suspect that similar polyglot language modeling is happenning in LLMs, such as ChatGPT and GPT-4, which we tested on our \dataset benchmark in Japanese, a language typologically distant from English.

\paragraph{Open-Domain and Multilingual Question Answering} 
\label{section:related_qa}
Much prior work proposed datasets for open-domain QA for English and beyond \cite{clark-etal-2020-tydi,asai2021xor,asai-etal-2022-mia,longpre2021mkqa,zhang-etal-2021-mr}.
Several works pointed out the problem of translation-based question answering evaluations \cite{clark-etal-2020-tydi,asai2021xor}: questions raised mainly by English speakers can diverge from information needs from speakers of other languages.
For instance, these translation-based benchmarks can overly represent English-centric topics, such
as American politics, sports, and culture.
To mitigate this English-centric problem, some datasets only sample questions from native speakers of each langauge \cite{clark-etal-2020-tydi}.
Consistent with such data creation methods, our \dataset consists of problems that are written by native Japanese speakers to evaluate  the qualifications and skills for medical practice in the country.

\section{Conclusion}
We presented our evaluations of the GPT APIs on the Japanese medical licensing examinations from 2018 to 2022.
The newest model, GPT-4, outperforms the others and manages to pass the examinations.
Through our benchmark, we highlighted several important limitations of the current LLM APIs when they are applied to a specialized domain in Japanese, a language typologically distant from English. 
We open-source our benchmark as \dataset, as well as the model outputs and meta information for future research.

\section*{Limitations}
This work evaluates large language models on Japanese medical licensing examinations.
We highlight several core limitations of our evaluations: \textbf{reproducibility and potential data leakage}, \textbf{language coverage}, and \textbf{scope of evaluation}.

First, as our experiments are performed using black-box LLM APIs, our results are not fully reproducibile, and the results may change with updates in the APIs.
Further, since the language model training data and setups are not well documented, there are potential risks of data leakage that overestimates the performance of LLMs.
To mitigate these issues,
we release all model outputs and experimental settings as well as the \dataset benchmark. 
This way if there is any update in the APIs, we can easily update our results and analyze changes in behaviors after the update.
\textbf{We have also included results from the current year (March 2023), which we believe is after the training of GPT-4, to address potential data leakage. We observed consistent performance with the previous years.}

Clearly, our benchmark is limited to the Japanese language and Japanese medical licensing examination.
It is an important research avenue to explore evaluations in more languages and domains.
Nonetheless, evaluation in the medical domain requires expertise, including knowledge specific to the country and its medical system and standard medical practice.
As discussed in this paper, there are potential risks if benchmarks are simply translated to various languages. 
The second author of this work is a doctor in a Japanese hospital, and such interdisciplinary efforts are necessary. 

Lastly, we note limitations in the scope of our evaluations.
For example, we did not use image information during evaluations because the current OpenAI LLM APIs do not support image input.
While some problems with images can be solved based solely on the problem text, many problems with images (and, of course, medical practice in general) need multimodal reasoning.
We leave it to future work to test models in multimodal settings.

Despite these challenges, we believe that it is important to benchmark black-box LLMs; they are increasingly used by people around the world across various disciplines.
We hope that our evaluations and \dataset benchmark will contribute to a better understanding of their behaviors, failures, and potential risks and benefits in diverse areas. 
\section*{In Memory of Professor Dragomir Radev}
The day after I completed this manuscript, I was eagerly awaiting your usual email and feedback.
To my great shock, I received the unexpected and sad news of your sudden passing.
I went to your office at Yale to leave white lilies, which I believe symbolize the purity of your life-long commitment to mentorship, education, and research.

LILY (Language, Information, and Learning at Yale) is also the name of your NLP lab at Yale University.
I am extremely fortunate to be part of the LILY lab since the beginning in Spring 2017.
I still vividly remember the day I visited your office for the first time.
At the time, I was in my senior year with almost no prior research experience.
Despite this, you kindly offered to mentor me on my research project.
After graduation, I sought your advice as to what I should do next.
Soon after, you and Professor Bob Frank from Yale Linguistics very kindly secured funding and offered me a research assistant position.
This experience became the foundation of my NLP research career.
During my Ph.D.\ at the University of Washington, we continued to meet regularly and collaborate on many exciting projects.

Among many other things, your attitude towards research has always struck me as passionate and open-minded.
The field of NLP has experienced many changes since I started.
We used to talk a lot about building core NLP models using LSTMs.
Now, we are seeing tremendous progress from large language models.
You always showed great enthusiasm for the latest advancements and how they are transforming the way we approach NLP problems.
Your passion for this field was contagious.
You consistently encouraged me to be open to new ideas, even when I was skeptical or anxious about new directions.
As you led by example, no matter how the field changes, researchers have a responsibility to demonstrate limitations and potentials of new technologies for society.

As one of the researchers who were extremely lucky to have you as an advisor, I feel obliged to pay it forward by continuing to support younger generations of scholars.
Thank you so much for the amazing six years.
May you find eternal rest and peace. \ja{ご冥福をお祈りいたします。} 

\hfill Jungo Kasai

\hfill \ja{笠井淳吾}

\section*{Acknowledgements}
We thank Noriyuki Kojima and Koji Shiono for their helpful feedback on this work.

\bibliography{custom}
\bibliographystyle{acl_natbib}

\clearpage
\appendix

\begin{appendices}
\section{Cateogries}
\label{appendix:categorization}
Our categorization of the exam problems is based on books widely used among Japanese medical students \cite{qb112,qb113,qb114,qb115,qb116}.
Table \ref{tab:category-translations} presents the Japanese-to-English translations of the category names.
We have 28 categories, ranging from public health to anesthesiology

\section{Additional Exam Information}
\label{appendix:data_statistics}
\begin{figure}[h]
\centering
\includegraphics[width=0.48\textwidth]{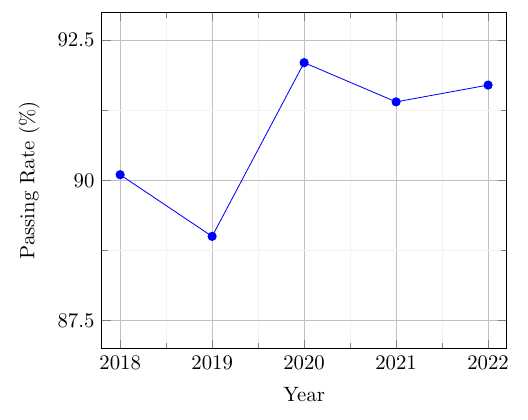}
\caption{Passing rates of Japanese national medical licensing examinations (NMPQE) over the past five years (2018-2022).}
\label{fig:passing_rates}
\end{figure}
\paragraph{Passing Rates}
Fig.\ \ref{fig:passing_rates} plots the passing rates of the Japanese medical licensing examination in the past five years.
The exam is typically taken by final-year medical students, and they obtain the Japanese medical license after passing the exam.
As shown in the figure, the passing rate slightly varies from year to year but are generally high (around 90\%).

\begin{figure*}[h]
\centering
\includegraphics[width=0.99\textwidth]{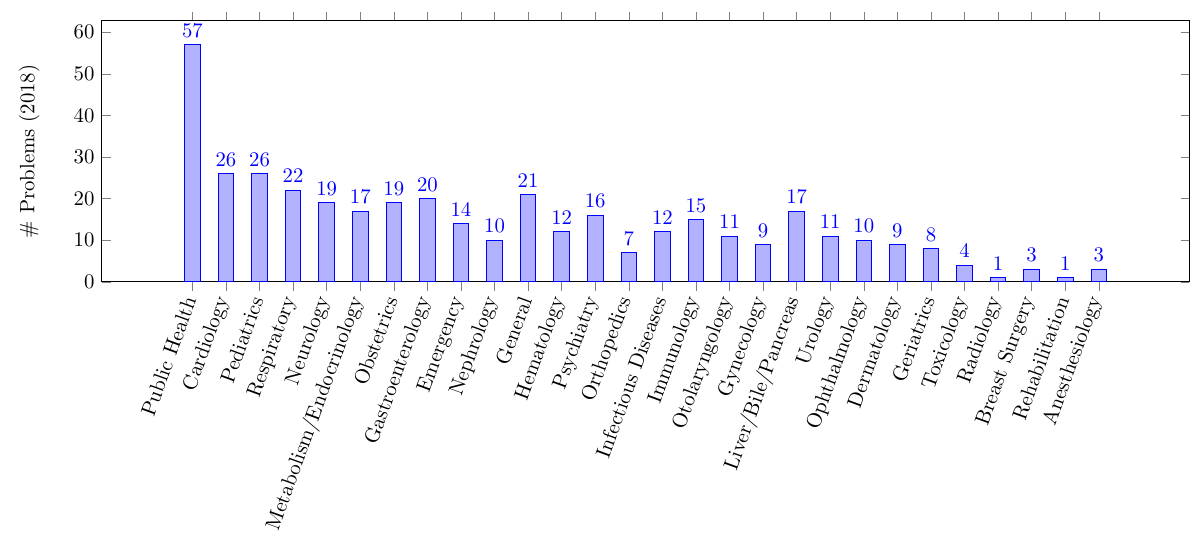}
\caption{Breakdown of the exam problems by category from the year 2018.}
\label{fig:cat_2018}
\end{figure*}

\begin{figure*}[h!]
\centering
\includegraphics[width=0.99\textwidth]{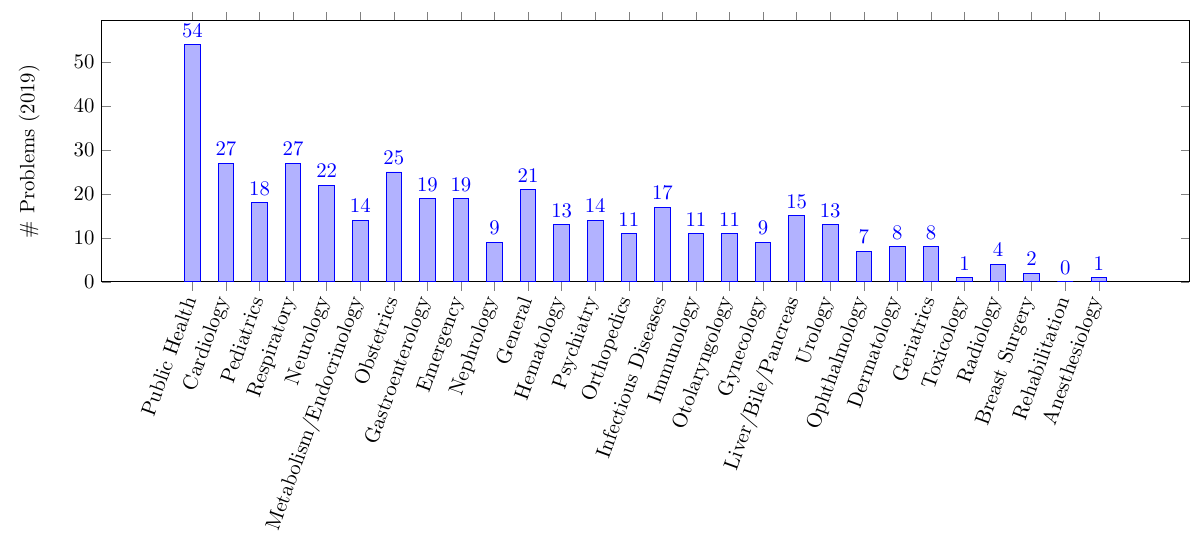}
\caption{Breakdown of the exam problems by category from the year 2019.}
\label{fig:cat_2019}
\end{figure*}

\begin{figure*}[h]
\centering
\includegraphics[width=0.99\textwidth]{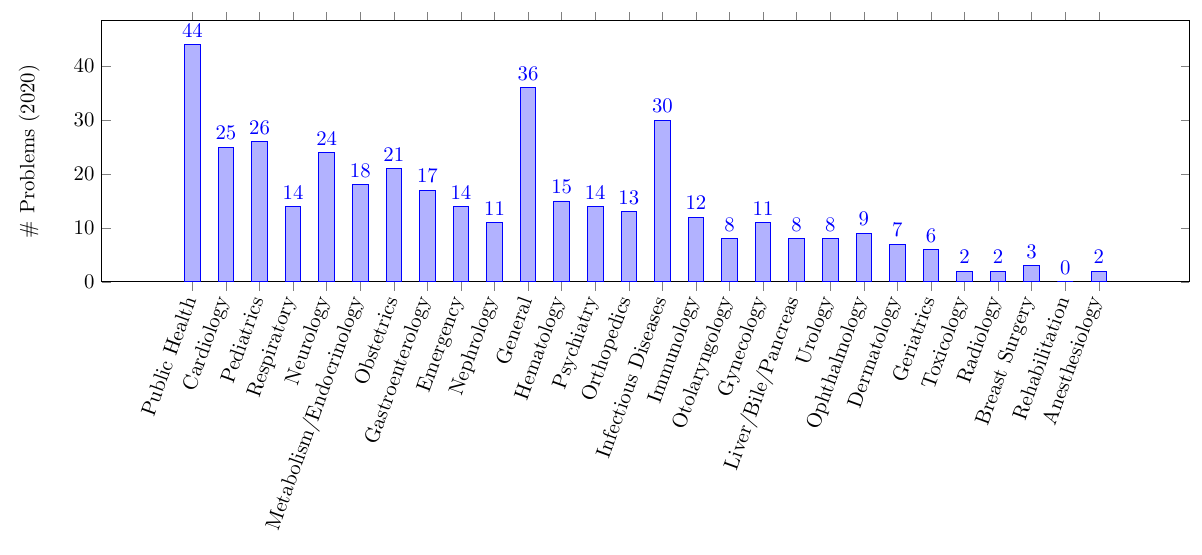}
\caption{Breakdown of the exam problems by category from the year 2020.}
\label{fig:cat_2020}
\end{figure*}

\begin{figure*}[h]
\centering
\includegraphics[width=0.99\textwidth]{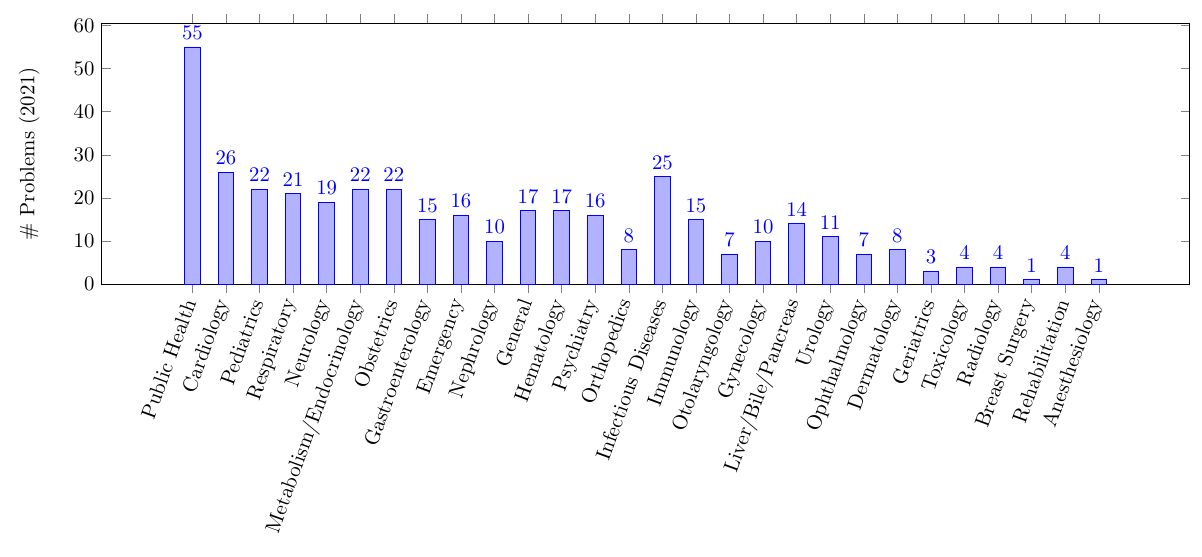}
\caption{Breakdown of the exam problems by category from the year 2021.}
\label{fig:cat_2021}
\end{figure*}

\paragraph{Breakdown by Category}
Figs.\ \ref{fig:cat_2018}-\ref{fig:cat_2021} plot the numbers of the problems over the 28 categories from year 2018 to 2021. 
The four years all have a catgory distribution similar to the exam in 2022 (Fig.\ \ref{fig:cat_2022}).
Different from the United States Medical Licensing Examination (USMLE), where three steps are taken over years, the Japanese medical licensing examination is usually taken once by final-year medical students. See also Fig.\ \ref{fig:Japan-vs-USA} for the standard timelines.

\begin{table*}[h]
\addtolength{\tabcolsep}{-2.5pt}
\small
\centering
\begin{tabular}{@{} l l  m{0.005em}  l l m{0.005em}  l l  @{}}
\toprule[.1em]
\textbf{Japanese} & \textbf{English}
&& \textbf{Japanese} & \textbf{English}
&& \textbf{Japanese} & \textbf{English} 
\\
\midrule[.1em]
\ja{公衆衛生} & Public Health
&& \ja{循環器} &  Cardiology 
&& \ja{小児科} & Pediatrics 
\\

\ja{呼吸器} & Respiratory
&& \ja{神経} & Neurology  
&& \ja{産科} & Obstetric
\\

\ja{代謝・内分泌} & Metabolism/Endocrinology
&& \ja{消化器} & Gastroenterology
&& \ja{救急} & Emergency
\\

\ja{腎臓} & Nephrology
&& \ja{総論} & General
&& \ja{血液} & Hematology
\\
\ja{精神科} & Psychiatry
&& \ja{整形外科} & Orthopedics 
&& \ja{感染症} & Infectious Diseases
\\

\ja{免疫} & Immunology
&& \ja{耳鼻科} & Otorhinolaryngology 
&& \ja{婦人科} & Gynecology
\\

\ja{肝・胆・膵} & Liver/Bile/Pancreas
&& \ja{泌尿器} & Urology
&& \ja{眼科} & Ophthalmology
\\

\ja{皮膚科} & Dermatology
&& \ja{老年医学} & Geriatrics
&& \ja{中毒} & Toxicology
\\

\ja{放射線} & Radiology
&& \ja{乳腺外科} & Breast Surgery
&& \ja{リハビリ} & Rehabilitation
\\
\ja{麻酔} & Anesthesiology
&&  &
&&  & 
\\

\bottomrule[.1em]
\end{tabular}
\caption{Translations of the 28 Japanese category names. The categorization is based on widely-used books for Japanese medical licensing exams \cite{qb112,qb113,qb114,qb115,qb116}.
The translations are done by the second author, a medical doctor based in Japan.
}
\label{tab:category-translations}
\end{table*}

\begin{figure*}[h]
\centering
\includegraphics[width=0.99\textwidth]{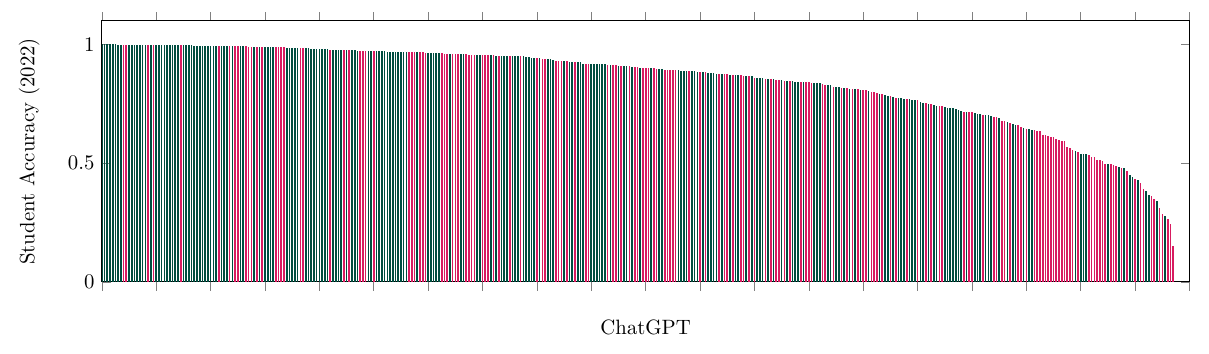}
\caption{Student (test taker) accuracy vs.\ ChatGPT results (\S\ref{sec:models_evals}). All problems from 2022 are sorted by the student accuracy, and the bar is \textcolor{ngreen}{green} when ChatGPT predicts the correct choice(s) and \textcolor{nred}{red} otherwise.
We see correlation between the student accuracy and the likelihood of the correct prediction.
}
\label{fig:116_student_chatgpt}
\end{figure*}

\begin{figure*}[h]
\centering
\includegraphics[width=0.99\textwidth]{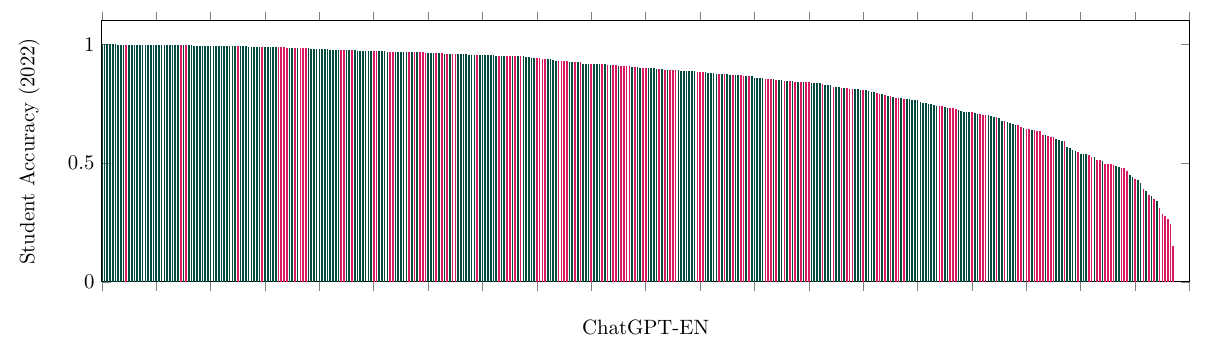}
\caption{Student (test taker) accuracy vs.\ ChatGPT-EN results (\S\ref{sec:models_evals}). All problems from 2022 are sorted by the student accuracy, and the bar is \textcolor{ngreen}{green} when ChatGPT-EN predicts the correct choice(s) and \textcolor{nred}{red} otherwise.
We see correlation between the student accuracy and the likelihood of the correct prediction.
}
\label{fig:116_student_chatgpt-en}
\end{figure*}

\begin{figure*}[h]
\centering
\includegraphics[width=0.99\textwidth]{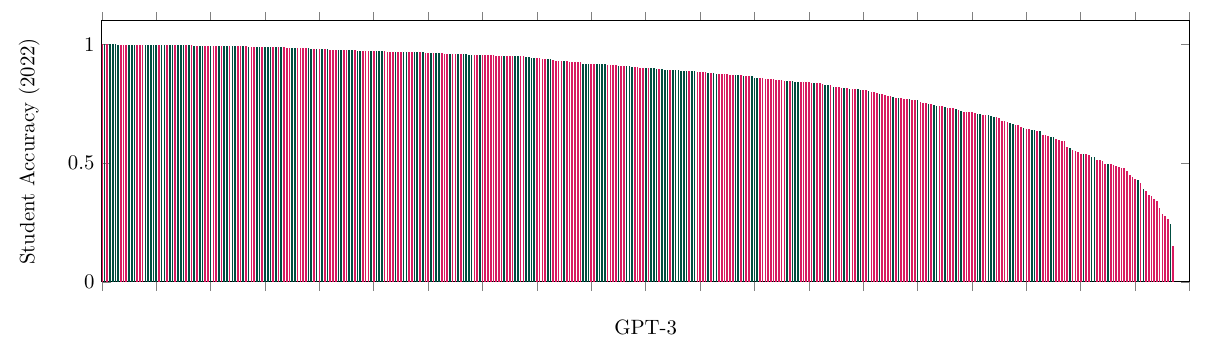}
\caption{Student (test taker) accuracy vs.\ GPT-3 results (\S\ref{sec:models_evals}). All problems from 2022 are sorted by the student accuracy, and the bar is \textcolor{ngreen}{green} when GPT-3 predicts the correct choice(s) and \textcolor{nred}{red} otherwise.
}
\label{fig:116_student_gpt3}
\end{figure*}

\section{Additional Settings and Results}
\label{appendix:setting}
\paragraph{GPT vs.\ Medical Students}
Similar to Fig.\ \ref{fig:116_student_gpts}, Figs.\ \ref{fig:116_student_chatgpt}-\ref{fig:116_student_gpt3} compare the student accuracy (the ratio of the students who selected the correct choice(s)) and the ChatGPT/ChatGPT-EN/GPT-3 results (\textcolor{ngreen}{green}: correct; \textcolor{nred}{red}: wrong)  for each problem from 2022.
We find correlation between the student accuracy and the likelihood of the correct prediction, suggesting that the LLMs struggle on questions that are also difficult for humans.
\paragraph{In-Context Examples}
Table \ref{tab:in-context_examples} our three in-context examples (translated into English for readability) for GPT-3 and ChatGPT/GPT-4. 
All of the three in-context examples were sampled from the Japanese medical licensing exam in 2006, which is also available on the official website of the Ministry of Health, Labour and Welfare of Japan (\ja{厚生労働省}).\footnote{\url{https://www.mhlw.go.jp/topics/2006/04/tp0419-1.html}.}
All our prompt templates are available online.\footnote{\url{https://github.com/jungokasai/IgakuQA}.}.

\begin{table*}
\small
\addtolength{\tabcolsep}{-3pt}
\begin{center}
\begin{tabular}{ L{7cm} L{7cm}}
\toprule
\textbf{Original (Japanese)} & \textbf{English}  \\

\midrule[.04em]

\ja{ヒトゲノム・遺伝子解析研究の被験者に対する説明で必須でないのはどれか。}
\par
\par
\ja{
a: 研究の目的
}
\par\ja{
b: 同意の自由
}
\par
\ja{
c: 匿名化の方法
}
\par
\ja{
d: 参加による不利益
}
\par
\ja{
e: 研究成果還元の保証
}
\par
\ja{
答え: e
}
&
Which of the following is not essential in explaining to
participants in a human genome/genetic analysis
study?
\par
\par
a: Purpose of the study
\par
b: Freedom to consent
\par
c: Method of anonymization
\par
d: Disadvantages of participation
\par
e: Assurance of research results sharing
\par
Answer: e
\\
\midrule[.1em]

\ja{57歳の男性。下水処理場のマンホール内で汚泥を外に搬出する作業を行っていたが、突然意識を失って倒れた。さらに救助しようとして中に入った同僚も急激に意識を失って倒れた。可能性が高いのはどれか。2つ選べ。}
\par
\par
\ja{
a: 酸素欠乏症
}
\par\ja{
b:  硫化水素中毒
}
\par
\ja{
c: 一酸化炭素中毒
}
\par
\ja{
d: 二酸化炭素中毒
}
\par
\ja{
e: 二酸化窒素中毒
}
\par
\ja{
答え: a, b
}
&
A 57-year-old man lost consciousness and collapsed while working to remove sludge from a manhole at a sewage treatment plant. A colleague who entered to assist also suddenly lost consciousness and collapsed. Which of the following is the most likely cause? Select two.
\par
\par
a: Oxygen deficiency
\par
b: Hydrogen sulfide poisoning
\par
c: Carbon monoxide poisoning
\par
d: Carbon dioxide poisoning
\par
e: Nitrogen dioxide poisoning
\par
Answer: a, b
\\
\midrule[.1em]

\ja{28歳の女性。妊娠30週。子宮底長は22cmで、腹部超音波検査で羊水はほとんど認めない。胎児で最も考えられるのはどれか。}
\par
\par
\ja{
a: 食道閉鎖
}
\par\ja{
b: 心室中隔欠損
}
\par
\ja{
c: 腎低形成
}
\par
\ja{
d: 鎖肛
}
\par
\ja{
e: 胎児水腫
}
\par
\ja{
答え: c
}
&
A 28-year-old woman at 30 weeks of gestation has a fundal height of 22 cm and almost no amniotic fluid is detected on abdominal ultrasound examination. What is the most likely condition in the fetus?
\par
\par
a: Esophageal atresia
\par
b: Ventricular septal defect
\par
c: Renal hypoplasia
\par
d: Anorectal malformation"
\par
e: Fetal hydrops
\par
Answer: c
\\

\bottomrule
\end{tabular}
\end{center}
\caption{
Our three in-context examples (translated into English for readability) for GPT-3 and ChatGPT/GPT-4. 
All of the three problems were sampled from the Japanese medical licensing exam in 2006.
}
\label{tab:in-context_examples}
\end{table*}

\end{appendices}

\end{document}